\documentclass{article}

    \PassOptionsToPackage{round, compress}{natbib}

\usepackage[final]{neurips_2024}

\usepackage[utf8]{inputenc} 
\usepackage[T1]{fontenc}    
\usepackage{hyperref}       
\usepackage{url}            
\usepackage{booktabs}       
\usepackage{amsfonts}       
\usepackage{nicefrac}       
\usepackage{microtype}      
\usepackage[usenames,dvipsnames]{xcolor} 

\usepackage{amsmath,amsfonts,bm}

\def\eqref#1{equation~\ref{#1}}

\def\1{\bm{1}}

\DeclareMathAlphabet{\mathsfit}{\encodingdefault}{\sfdefault}{m}{sl}
\SetMathAlphabet{\mathsfit}{bold}{\encodingdefault}{\sfdefault}{bx}{n}

\newcommand{\pdata}{p_{\rm{data}}}

\newcommand{\E}{\mathbb{E}}

\newcommand{\R}{\mathbb{R}}

\newcommand{\KL}{D_{\mathrm{KL}}}

\usepackage{natbib} 
\usepackage{algorithm}
\usepackage{algorithmic}
\usepackage[capitalize]{cleveref}
\usepackage{dsfont}

\usepackage{microtype}
\usepackage{graphicx}
\usepackage{subfigure}
\usepackage{booktabs} 
\usepackage{subcaption}
\usepackage{dsfont}
\usepackage{multirow}
\usepackage{caption}
\usepackage{tikz}
\usetikzlibrary{positioning, fit, calc, arrows.meta, shapes}
\usetikzlibrary{automata,arrows,positioning,calc}
\usetikzlibrary{shapes.multipart}
\usepackage{wrapfig}
\usetikzlibrary{positioning, arrows.meta, fit, backgrounds, decorations.pathreplacing}

\renewcommand{\pdata}{p_{\mathcal{D}}}
\usepackage{relsize}
\definecolor{customlinkcolor}{HTML}{2774AE} 
\definecolor{customcitecolor}{HTML}{2774AE} 
\hypersetup{
    colorlinks=true,
    linkcolor=customlinkcolor,
    anchorcolor=black, 
    citecolor=customlinkcolor,
    urlcolor=customlinkcolor 
}

\makeatletter
\renewcommand\@fnsymbol[1]{}
\makeatother

\title{Latent Plan Transformer for Trajectory Abstraction: Planning as Latent Space Inference}

\author{%
  Deqian Kong$^{1,\star}$\thanks{$^\star$Equal Contribution.}\thanks{Project page: \href{https://sites.google.com/view/latent-plan-transformer}{https://sites.google.com/view/latent-plan-transformer}.}\thanks{Code: \href{https://github.com/mingluzhao/Latent-Plan-Transformer}{https://github.com/mingluzhao/Latent-Plan-Transformer}.}, Dehong Xu$^{1,\star}$, Minglu Zhao$^{1,\star}$, Bo Pang$^2$, Jianwen Xie$^3$, \\
\textbf{Andrew Lizarraga$^1$, {Yuhao Huang}$^4$, {Sirui Xie}$^{5,\star}$, {Ying Nian Wu}$^1$}\\
\\
$^1$Department of Statistics and Data Science, UCLA\\
$^2$Salesforce Research $^3$Akool Research
$^4$Xi'an Jiaotong University\\
$^5$Department of Computer Science, UCLA \\
}

\begin{document}

\maketitle

\begin{abstract}
In tasks aiming for long-term returns, planning becomes essential. We study generative modeling for planning with datasets repurposed from offline reinforcement learning. Specifically, we identify temporal consistency in the absence of step-wise rewards as one key technical challenge. We introduce the Latent Plan Transformer (LPT), a novel model that leverages a latent variable to connect a Transformer-based trajectory generator and the final return. LPT can be learned with maximum likelihood estimation on trajectory-return pairs. In learning, posterior sampling of the latent variable naturally integrates sub-trajectories to form a consistent abstraction despite the finite context. At test time, the latent variable is inferred from an expected return before policy execution, realizing the idea of \textit{planning as inference}.  
Our experiments demonstrate that LPT can discover improved decisions from suboptimal trajectories, achieving competitive performance across several benchmarks, including Gym-Mujoco, Franka Kitchen, Maze2D, and Connect Four. It exhibits capabilities in nuanced credit assignments, trajectory stitching, and adaptation to environmental contingencies. These results validate that latent variable inference can be a strong alternative to step-wise reward prompting. 
\end{abstract}

\section{Introduction}
\label{sec:intro}
Decision Transformer (DT)~\citep{chen2021decision} and some concurrent work~\citep{janner2021offline} have popularized the research agenda of decision-making via generative modeling. The general idea is to consider decision-making as a generative process that takes in a representation of the task objective, e.g. the rewards or returns of a trajectory, and outputs a representation of the trajectory. Intuitively, a purposeful decision-making process should shift the trajectory distribution towards regimes with higher returns. In the classical decision-making literature, this is achieved by two interweaving processes, {policy evaluation} and {policy improvement} \citep{sutton2018reinforcement}. {Policy evaluation} promotes consistency in the estimated correlations between the trajectories and the returns. In DT, this is realized by the maximum likelihood estimation (MLE) of the joint distribution of sequences consisting of states, actions, and return-to-gos (RTG). {Policy improvement} shifts the distribution to improve the status quo expectation of the returns. In DT, this is naturally entailed since the policy is a distribution of actions conditioned on step-wise RTGs. 

In this work, we are interested in the problem of \textit{planning}. Among various ways to identify \textit{planning} as a special class of decision-making problems, we pay particular attention to its data specification and inductive biases. As designing step-wise rewards requires significant effort and domain expertise, we focus on the problem of learning from trajectory-return pairs, where a trajectory is a sequence of states and actions, and the return is its total rewards. 
This design choice forces the agents to predict into the long-term future and figure out step-wise credits by themselves. A competitive Temporal Difference (TD) learning baseline, CQL~\citep{kumar2020conservative}, was reported to be fragile under this data specification~\citep{chen2021decision}. 

Our design of inductive biases reflects our intuition of a \textit{plan}. While a policy is a factor of the trajectory distribution, a \textit{plan} is an abstraction lifted from the space of trajectories. As a plan is always made in advance of receiving returns, it implies \textit{significance}, \textit{persistence}, and \textit{contingency}. An agent should plan for more significant returns. It should be persistent in its plan even if the return is assigned in hindsight. It should also be adaptable to the environment's changes during the execution of the plan. We formulate this hierarchy of decision-making with a top-down latent variable model. The latent variable we introduce is effectively a \textit{plan}, for it decouples the trajectory generation from the expected improvement of returns. The autoregressive policy always consults this temporally extended latent variable to be persistent in the plan. The top-down structure enables the agent to disentangle the variations in its plan from the environment's contingencies.  

In this work, we introduce the Latent Plan Transformer~(LPT), a novel generative model featuring a latent vector modeled by a neural transformation of Gaussian white noise, a Transformer-based policy conditioned on this latent vector and a return estimation model. LPT is learned by maximum likelihood estimation (MLE). Given an expected return, posterior inference of the latent vector in LPT is an explicit process for iterative refinement of the \textit{plan}. The inferred latent variable replaces RTG in the conditioning of the auto-regressive policy, providing richer information about the anticipated future. 
We further develop a mode-seeking sampling scheme that strongly enforce the temporal consistency for long-range planning, which is particularly effective in \textit{stitch} trajectory, {i.e.}, to compose parts of sub-optimal trajectories to reach far beyond~\citep{fu2020d4rl}. LPT demonstrates competitive performance in Gym-Mujoco locomotion, Franka kitchen, goal-reaching tasks in maze2d and antmaze, and a contingent planning task Connect Four. These empirical results support that latent variable inference can enable and improve planning in the absence of step-wise rewards. 

\section{Background}
\label{sec:problem}
A sequential decision-making problem can be formulated with a decision process $\langle S, A, H, Tr, r, \rho\rangle$ that contains a set $S$ of states and a set $A$ of actions. Horizon $H$ is the maximum number of steps the agent can execute before the termination of the sequence. We further employ $S^{+}$ to denote the set of all non-empty state sequences within the horizon and $A^{+}$ for action sequences likewise. $Tr: S^{+}\times A^{+}\mapsto \Pi(S)$ is the transition that returns a distribution over the next state. $r: S^{+}\times A^{+}\mapsto\R$ specifies the real-valued reward at each step. $\rho: \Pi(S)$ is the initial state distribution that is always uncontrollable to the agent. The agent's decisions follow a policy $\pi: S^{+}\times A^{+}\mapsto\Pi(A)$. In each episode, the agent interacts with the transition model to generate a trajectory $\tau= (s_1,a_1,s_2,a_2 ...,s_H,a_H)$. 

The objective of sequential decision-making is typically formulated as the expected trajectory return $y=\sum_{t=0}^H r_t$, $Q=\mathbb{E}_{p(\tau)}[y]$. Conventional RL algorithms solve for a policy $\pi(a_t|s_t, {\ast})$, where the conditioning $\ast$ denotes the optimal expected return. DT generalizes this policy to $\pi(a_t|s_{\leq t}, a_{<t}, RTG_{\leq t})$, by fitting the joint distribution $p(s_1, a_1, RTG_1, ... s_T, a_T, RTG_T)$ with a Transformer. $RTG_t$ is the return-to-go from step $t$ to the horizon $H$, $RTG_t=\sum_{k=t}^{H}r(s_{\leq k}, a_{\leq k})$.
It is a useful indication of future rewards, especially when rewards are dense and informative. 

However, $RTG$ becomes less reliable when rewards are sparse or have non-trivial relations with the return. Distributing the return to each step is a credit assignment problem. Consider an example of an ideal credit assignment mechanism: When students receive partial credits for their incomplete answers, it's more fair to give points equal to the full marks minus the expected points for all possible ways to finish the answer, rather than assuming students have no knowledge of the remaining parts.
This credit assignment mechanism can be formalized as, $RTG_t^Q = \sum_{k=t}^K r(s_{\leq k}, a_{\leq k}) + \mathbb{E}[Q(s_{\leq K}, a_{\leq K})]$.
Here $Q$ can be estimated using deep TD learning with multi-step returns. \citet{yamagata2023q} instantiate a Markovian version and demonstrate improvement in trajectory \textit{stiching}. 

Whatever credit assignment we use, be it $RTG$ or $RTG^Q$, the purpose is to explicitly model the statistical association between trajectory steps and final returns. This effort is believed to be necessary because of the exponential complexity of the trajectory space. This belief, however, can be re-examined given the success of sequence modeling. We explore an alternative design choice by directly associating the latent vector that generates the trajectory with the return.

\section{Latent Plan Transformer (LPT)}
\label{sec:method}
\subsection{Model}
\label{sec:model}
\begin{figure}[h]
\centering
\resizebox{0.55\linewidth}{!}{
\begin{tikzpicture}[->, >=stealth', auto, thick, node distance=1.3cm]
    \tikzstyle{every state}=[fill=white,draw=black,thick,text=black,scale=1,minimum size=0.7cm]
    \node[state]    (z)                                 {$z$};
    \node[state]    (z0)[above =0.6cm of z,label=center:$z_0$] {\phantom{$z$}};
    \node[state]    (x)[below left=0.48cm and 0.36cm of z,fill=black!20,label=center:$\tau$] {\phantom{$z$}};
    \node[state]    (y)[below right=0.48cm and 0.36cm of z,fill=black!20,label=center:$y$] {\phantom{$z$}};
    \path
    (z) edge[right]               (y)
    (z) edge[left]                (x)
    (z0) edge[left]      node{\small $z=U_\alpha{(z_0)}$}          (z);
    \node       (pa)[right= -0.05cm of z]    {\small $p_\alpha(z)$};
    \node       (px)[below= -0.08cm of x]    {\small $p_\beta(\tau|z)$};
    \node       (py)[below= -0.08cm of y]    {\small $p_\gamma(y|z)$};

    \draw[dashed, -] (1.75cm, -1.8cm) -- (1.75cm, 2.2cm);
    
    \begin{scope}[shift={(4.5cm,1cm)}, block/.style={rectangle, draw, fill=black!10, minimum width=3cm, minimum height=0.6cm, align=center, rounded corners},
    arrow/.style={thin, -stealth'},
    bigbox/.style={draw, thick, rounded corners, inner sep=5pt, dashed}]
    
        \node[block] (cross) {\small Cross-attention};
        \node[block, below=0.75cm of cross] (causal) {\small Causal Transformer};

        \begin{scope}[on background layer]
        \node[bigbox, fit=(cross) (causal)] (mainbox) {};
        \end{scope}

        \node[above right=0.1cm and -0.5cm of cross.north east, anchor=south west] (N) {$\times N$};

        \node[left=0.5cm of mainbox.west] (z2) {$z$};
        \node[below=0.3 cm of mainbox] (xi) {\small$\tau=(s_1,a_1,s_2,a_2,\dots, s_H, a_H)$};
        \node[above=0.3 cm of mainbox] (xo) {\small$a_1,a_2,\dots,\cdots,a_H$};

        \draw[arrow] (z2) -| ($(cross.south west)!0.25!(cross.south east)$);
        \draw[arrow] (z2) -| ($(cross.south west)!0.5!(cross.south east)$);
        \draw[arrow] (xi) -- (causal);
        \draw[arrow] (cross) -- (xo);

        \draw[arrow] ($(causal.north west)!0.75!(causal.north east)$) -- ($(cross.south west)!0.75!(cross.south east)$);
    \end{scope}
\end{tikzpicture}
}
\caption{\textit{Left}: Overview of Latent Plan Transformer (LPT). $z\in\mathbb{R}^d$ is the latent vector. The prior distribution of $z$ is a neural transformation of $z_0$, i.e., $z = U_\alpha(z_0)$, $z_0 \sim {\cal N}(0, I_d)$. Given $z$, $\tau$ and $y$ are independent. $p_\beta(\tau|z)$ is the trajectory generator. $p_\gamma(y|z)$ is the return predictor.  \textit{Right}: Illustration of trajectory generator $p_\beta(\tau|z)$. }
\label{fig:LPT}
\end{figure}
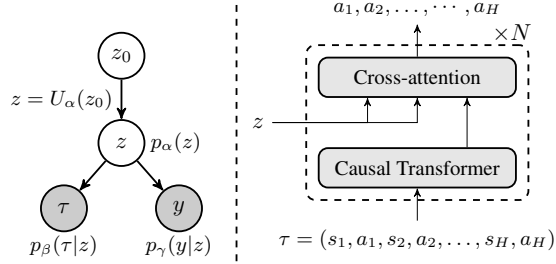  

Given a variable-length trajectory $\tau$, $z \in \mathbb{R}^d$ is a vector that represents $\tau$ in the latent space. $y \in \mathbb{R}$ is the return of the trajectory. The joint distribution of the trajectory and its return is defined as $p(\tau,y)$.

The latent trajectory variable $z$, conceptualized as a plan, is posited to decouple the autoregressive policy and return estimation.
From a statistical standpoint, with $z$ given, we assume that $\tau$ and $y$ are conditionally independent, positioning $z$ as the information bottleneck. 
Under this assumption, the Latent Plan Transformer (LPT) can be defined as,
\begin{align}
p_{\theta}(\tau, y, z) &= p_{\alpha}(z) p_\beta(\tau|z) p_\gamma(y|z),
\label{eq:joint}
\end{align}
where $\theta = (\alpha, \beta, \gamma)$. 
LPT approximates the data distribution $p_\mathrm{data}(\tau, y)$ using the marginal distribution $p_\theta(\tau, y) = \int p_\theta(\tau, y, z) dz$. It also establishes a generation process,
\begin{align}
z \sim p_{\alpha}(z), \quad [\tau|z] \sim p_\beta(\tau|z), \quad [y|z] \sim p_\gamma(y|z).
\end{align}

The prior model $p_{\alpha}(z)$ is an implicit generator, defined as a learnable neural transformation of an isotropic Gaussian, $z = U_\alpha(z_0)$ and $z_0\sim \mathcal{N}(0,I_d)$.
$U_\alpha(\cdot)$ is an expressive neural network, such as the UNet~\citep{ronneberger2015u}. This approach is inspired by, yet contrasts with~\citet{pang2020learning}, wherein the latent space prior is modeled as an Energy-based Model (EBM) \citep{xie2016theory}. While EBM offers explicit unnormalized density, its sampling process is complex. Conversely, our model provides an implicit density with simpler sampling.

The trajectory generator $p_{\beta}(\tau|z)$ is a conditional autoregressive model with finite context $K$, $p_\beta(\tau|z) = \prod_{t=1}^H p_\beta(\tau_{(t)}|\tau_{(t-K)}, ..., \tau_{(t-1)}, z)$
where $\tau_{(t)}=(s_t, a_t)$. It can be parameterized by a causal Transformer with parameter $\beta$, similar to Decision Transformer~\citep{chen2021decision}. Specifically, the latent variable $z$ is included in trajectory generation using cross-attention, as shown in~\cref{fig:LPT} and controls each step of the autoregressive trajectory generation as $p_\beta(a_t|s_{t-K:t},a_{t-K:t-1}, z)$. The action is assumed to follow a single-mode Gaussian distribution, i.e. $a_t\sim \mathcal{N}(g_\beta(s_{t-K:t},a_{t-K:t-1}, z), I_{|A|})$. 

The return predictor is a non-linear regression on the latent trajectory variable $z$, modeled as $p_{\gamma}(y|z) = \mathcal{N}(r_\gamma(z), \sigma^2)$.
It directly predicts the final return from the latent variable $z$.
The function $r_\gamma(z)$ is a small multi-layer perceptron (MLP) that estimates $y$ based on $z$. The variance $\sigma^2$, is treated as the hyper-parameter in our setting. 

\subsection{Offline Learning}
\label{sec:offline_learning}

With a set of offline training examples $\{(\tau_i, y_i)\}_{i=1}^n$, we aim to learn Latent Plan Transformer (LPT) through maximum likelihood estimation (MLE). The log-likelihood function is defined as $L( \theta) = \sum_{i=1}^{n} \log p_\theta(\tau_i, y_i)$.
The joint probability of the trajectory and final return is
\begin{align}
   p_\theta(\tau, y) = \int p_{\beta}(\tau|z = U_\alpha(z_0)) p_{\gamma}(y| z = U_\alpha(z_0)) p_0(z_0) dz_0,
\end{align}
where $p_0(z_0)={\cal N}(0, I_d)$.
The learning gradient of log-likelihood can be calculated according to
\begin{align} 
\begin{split}
    \nabla_\theta  \log p_\theta(\tau, y)&=
    \E_{p_\theta(z_0|\tau, y)} [\nabla_\theta\log p_\beta(\tau|U_\alpha(z_0)) + \nabla_\theta\log p_\gamma(y|U_\alpha(z_0))].  
\label{eq:grad}
\end{split}
\end{align}
The full derivation of the learning method is in~\cref{appen:model}.
Let $\delta_\alpha, \delta_\beta, \delta_\gamma$ represent the expected gradients of $L(\theta)$ with respect to the model parameters $\alpha,\beta,\gamma$, respectively. The learning gradients for each component are formulated as follows.

For the prior model $p_\alpha(z)$,
\begin{align} 
  \delta_\alpha(\tau, y) &= \E_{p_\theta(z_0|\tau, y)} [\nabla_\alpha(\log p_\beta(\tau|z=U_\alpha(z_0))+ \nabla_\alpha\log p_\gamma(y|z=U_\alpha(z_0))]. \nonumber
\end{align}
For the trajectory generator,
\begin{align} 
  \delta_\beta(\tau, y) = \E_{p_\theta(z_0|\tau, y)} [\nabla_\beta \log p_{\beta}(\tau|z=U_\alpha(z_0))], \nonumber
\end{align} 
For the return predictor,
\begin{align} 
  \delta_\gamma(\tau, y) = \E_{p_\theta(z_0|\tau, y)} [\nabla_\gamma \log p_{\gamma}(y|z=U_\alpha(z_0))]. \nonumber
\end{align} 

Estimating these expectations requires Markov Chain Monte Carlo~(MCMC) sampling of the posterior distribution $p_\theta(z_0|\tau,y)$. We use the Langevin dynamics~\citep{neal2011mcmc} for MCMC sampling, iterating as follows for a target distribution $\pi(z)$:
\begin{align} 
z^{k+1} = z^k + s \nabla_z \log \pi(z^k) + \sqrt{2s} \epsilon^k, 
 \label{eq:Langevin}
\end{align}
where $k$ indexes the time step of the Langevin dynamics, $s$ is the step size, and $\epsilon^k \sim {\mathcal N}(0, I_d)$ is the Gaussian white noise. Here, $\pi(z)$ is instantiated as the posterior distribution $p_\theta(z_0|\tau, y)$. We have $p_\theta(z_0|\tau, y)\propto p_0(z_0)p_\gamma(y|z)p_\beta(\tau|z)$, where $z=U_\alpha(z_0)$, such that the gradient is
\begin{align}
    \nabla_{z_0} \log p_\theta(z_0|\tau, y) 
    =\nabla_{z_0}\underbrace{\log p_0(z_0)}_{\text{prior}}+\nabla_{z_0}\underbrace{\log p_\gamma(y|z)}_{\text{return prediction}}+\underbrace{\sum\nolimits_{t=1}^H\nabla_{z_0}\log p_\beta(\tau_{(t)}|\tau_{(t-K:t-1)},z)}_{\text{aggregating finite-context sub-trajectories}}. \nonumber
\end{align}
This demonstrates that the posterior inference of $z$ is an explicit process of optimizing a plan given its likelihood. In the presence of a finite context, $p_\beta(\tau|z)$ parametrized with Transformer can only account for sub-trajectories with a maximum length of $K$. The latent variable $z$ serves as an abstraction that integrates information from both the final return and sub-trajectories using gradients.

The sampling process starts by initializing $z_0^{k=0}$ from a standard normal distribution ${\mathcal N}(0, I_d)$. We then apply $N$ steps of Langevin dynamics (e.g., $N=15$) to approximate the posterior distribution, making our learning algorithm an approximate MLE. For a theoretical understanding of this noise-initialized finite-step MCMC, see \citet{pang2020learning,nijkamp2020learning2,XieZXL023}. 
However, for large horizons (e.g.,$H$=1000), this method becomes slow and memory-intensive. To mitigate this, we adopt the persistent Markov Chain (PMC)~\citep{tieleman2008training,xie2016theory,han2017abp}, which amortizes sampling across training iterations. During training, $z_0^{k=0}$ is initialized from the previous iteration and the number of updates is reduced to $N=2$ steps. See~\cref{appen:training} for training and architecture details.

\subsection{Planning as Inference}
\label{sec:inference}
The MLE learning of LPT gives us an agent that can plan. During testing, we first infer the latent $z_0$ given the desired return $y$ using Bayes' rule,
\begin{align}
    z_0\sim p_\theta(z_0|y) \propto p_0(z_0)p_\gamma(y|z=U_\alpha(z_0)).
\label{eq:p_z0_y}
\end{align}
This posterior sampling is achieved using Langevin dynamics similar to the training process. Specifically, we replace the target distribution in~\cref{eq:Langevin} with $p_\theta(z_0|y)$ and run MCMC for a fixed number of steps. Sampling from $p_\theta(z_0|y)$ eliminates the need for expensive back-propagation through the trajectory generator $p_\beta(\tau|z)$. 

This posterior sampling of $p(z_0|y)$ is an explicit process that iteratively refines the latent plan $z$, increasing its likelihood given the desired final return. It aligns with our intuition that planning is an inference process. This inferred $z$, fixed ahead of the policy execution, effectively serves as a plan. At each step, the agent consults this plan to generate actions conditioned on the current state and recent history, $a_t \sim p_\beta(a_t|s_{t-K:t-1},a_{t-K:t-1}, z=U_\alpha(z_0)).$

Once a decision is made, the environment's (possibly non-Markovian) transition $s_{t+1}\sim p(s_{t+1}|a_t, s_{t})$ emits the next state. This sequential decision-making process iterates the sampling of $s_t$ and $a_t$ until termination at the horizon.

\paragraph{Exploitation-inclined Inference (EI)}
Inspired by the classifier guidance (CG)~\citep{dhariwal2021diffusion,ho2022classifier} in conditional diffusion models, we introduce a guidance weight $w$ to the original posterior in \cref{eq:p_z0_y} 
\begin{align}
\label{eq:cg}
\Tilde{p}_\theta(z_0|y)\propto p_0(z_0)p_\gamma(y|z)^w, z=U_\alpha(z_0),
\end{align}
which has the score $\nabla_{z_0} \log \Tilde{p}_\theta(z_0|y)=\nabla_{z_0}{\log p_0(z_0)}+w\nabla_{z_0}{\log p_\gamma(y|z)}$.
This guidance weight $w$ controls the interpolation between exploration and exploitation. When $w=1$, the sampled plans collectively represent the posterior density and account for Bayesian uncertainty, resulting in a provably efficient exploration scheme~\citep{osband2017posterior}. When $w>1$, the sampled plans are more concentrated around the modes of the posterior distribution, which are plans more likely to the agent. The larger the value of $w$, the more confident the agent becomes, and the stronger the inclination towards exploitation.

An overview of the algorithms for both offline learning and inference can be found in the following.

\begin{algorithm}[h]
\caption{Offline learning}
\label{algo:learning}
\begin{algorithmic}
    \STATE {\bfseries Input:} Learning iterations~$T$, initial parameters~$\theta_0 = (\alpha_0, \beta_0, \gamma_0)$, offline training samples~$\mathcal{D}=\{\tau_i, y_i\}_{i=1}^n$, posterior sampling step size $s$, the number of steps $N$, and the learning rate $\eta_0,\eta_1,\eta_2$.
    \STATE {\bfseries Output:} $\theta_T$
    \FOR{$t=1$ {\bfseries to} $T$}
    \STATE{1.\textbf{Posterior sampling}}: 
    For each $(\tau_i, y_i)$, sample $z_0 \sim {p}_{\theta_t}(z_0|\tau_i, y_i)$ using~\cref{eq:Langevin} with $N$ steps and step-size $s$, where the target distribution $\pi$ is  ${p}_{\theta_t}(z_0|\tau_i, y_i)$. 
    \STATE{2.\textbf{Learn prior model} $p_\alpha(z)$, \textbf{trajectory generator} $p_\beta(\tau|z)$ and \textbf{return predictor} $p_\gamma(y|z)$}: \\
    $\alpha_{t+1} = \alpha_t + \eta_0 \frac1n\sum_i\delta_\alpha(\tau_i,y_i)$;
    $\beta_{t+1} = \beta_t + \eta_1 \frac{1}{n} \sum_{i}\delta_\beta(\tau_i,y_i)$;
    $\gamma_{t+1} = \gamma_t + \eta_2 \frac{1}{n} \sum_{i}\delta_\gamma(\tau_i,y_i)$ as in~\cref{sec:offline_learning}.
    \ENDFOR
\end{algorithmic}
\end{algorithm}
\begin{algorithm}[h]
\caption{Planning as inference}
\label{algo:online}
\begin{algorithmic}
    \STATE {\bfseries Input:} Expected return $y$, a trained model on offline dataset $\theta$, posterior sampling step size $s$ and the number of steps $N$, Horizon $H$ and an evaluation environment.
    \STATE {\bfseries Output:} Trajectory $\tau$
    \IF{Exploitation-inclined Inference (EI)}
    \STATE{Sample $z_0\sim \Tilde{p}_\theta(z_0|y)$ as in~\cref{eq:cg} using~\cref{eq:Langevin} with $N$ steps and step size $s$, where the target distribution $\pi$ is replaced by $\Tilde{p}_\theta(z_0|y)\propto p_0(z_0)p_\gamma(y|z=U_\alpha(z_0))^w$ and $z=U_\alpha(z_0)$.}
    \ELSE
    \STATE{Sample $z_0\sim p_\theta(z_0|y)$ as in~\cref{eq:p_z0_y} using~\cref{eq:Langevin} with $N$ steps and step size $s$, where $\pi$ is replaced by $p_\theta(z_0|y)\propto p_0(z_0)p_\gamma(y|z=U_\alpha(z_0))$ and $z=U_\alpha(z_0)$.}
    \ENDIF
    \WHILE{current time step $t \le H$}
    \STATE Sample $a_t$ using trajectory generator as $a_t \sim p_\beta(a_t|s_{t-K:t-1},a_{t-K:t-1}, z=U_\alpha(z_0))$. \\Once a decision is made, the environment's transition $s_{t+1}\sim p(s_{t+1}|a_t, s_{t})$ emits the next state.\\
    \ENDWHILE
\end{algorithmic}
\end{algorithm}

\section{A Sequential Decision-Making Perspective}
\label{sec:analysis}
We approach the sequential decision-making problem with techniques from generative modeling. In particular, our data specification of trajectory-return pairs omits step-wise rewards, based on the belief that the step-wise reward function is only a proxy of the trajectory return. However, step-wise rewards are indispensable input to classical decision-making algorithms. Accumulating the rewards from the current step to the future gives us the $RTG$, which naturally hints the future progress of the trajectory. How is temporal consistency enforced in our model without the assistance of the $RTG$s?

Without loss of generality, consider the trajectory distribution conditioned on a single return value $y$. The MLE objective is equivalent to minimizing the KL divergence between the data distribution and model distribution,
$\KL(\pdata^y(\tau)\| p_{\theta}^y(\tau))$. Here, $\pdata$ denotes the data distribution and $p_\theta$ denotes the model distribution. 
MLE upon autoregressive modeling imposes additional inductive biases by transforming the objective to $\KL(p_{\mathcal{D},\text{AR}}^y(\tau)\| p_{\theta, \text{AR}}^y(\tau))$, which is reduced to next-token prediction for behavior cloning and transition model estimation: 
\begin{align*}  \sum_{t=1}^{H}\!\underbrace{\KL(\pdata^y(a_t|s_{1:t},\!a_{1:t-1})\|p_{\theta}^y(a_t|s_{1:t},\!a_{1:t-1}))}_{\text{behavior cloning}}
\!+\!\sum_{t=1}^{H}\!\underbrace{\KL(\pdata^y(s_{t+1}|s_{1:t},\!a_{1:t})\| p_{\theta}^y(s_{t+1}|s_{1:t},\!a_{1:t}))}_{\text{transition model estimation}}.
\end{align*}
However, behavior cloning is believed to suffer from drifting errors since it ignores \textit{covariate shifts} in future steps~\citep{ross2010efficient}. This concern is unique to sequential decision-making, as the agent cannot control the next state from a stochastic environment, like generating the next text token. 

This temporal consistency issue could be alleviated by additionally modeling the sequence of $RTG$. Denote $\rho = (RTG_0, RTG_1, ... RTG_H)$. Modeling the joint distribution is to minimize
\begin{equation}
\label{eq:rtg}
\begin{split}
    &\KL(\pdata^y(\tau, \rho)\| p_{\theta}^y(\tau, \rho))=\KL(\pdata^y(\tau)\| p_{\theta}^y(\tau))+\KL(\pdata^y(\rho|\tau)\| p_{\theta}^y(\rho|\tau)) \\
    =&\KL(p_{\mathcal{D}, \text{AR}}^y(\tau)\| p_{\theta, \text{AR}}^y(\tau))+\E_{\pdata^y(\tau)}[\sum\nolimits_{t=1}^{H}\underbrace{\KL(\pdata^y(RTG_t|\tau)\| p_{\theta}^y(RTG_t|\tau))}_{\text{RTG prediction}}].
\end{split}
\end{equation}
Note that the \emph{RTG prediction} term is conditioned on the entire trajectory, including the future steps. Minimizing this additional KL divergence correlates predicted $RTG$s with hindsight trajectory-to-go. 

Our modeling of the latent trajectory variable $z$ provides an alternative solution to the temporal consistency issue. \cref{eq:grad} is minimizing the KL divergence
\begin{equation}
\label{eq:plan}
\begin{split}
    &\KL(\pdata^y(\tau, z)\| p_{\theta}^y(\tau, z))=\KL(\pdata^y(\tau)\| p_{\theta}^y(\tau))+\KL(p_{\bar{\theta}}^y(z|\tau)\| p_{\theta}^y(z|\tau))\\
    =&\KL(p_{\mathcal{D}, \text{AR}}^y(\tau)\| p_{\theta, \text{AR}}^y(\tau)) +\E_{\pdata^y(\tau)}[\underbrace{\KL(p_{\bar{\theta}}^y(z|\tau)\| p_{\theta}^y(z|\tau))}_{\text{plan prediction}}],
\end{split}
\end{equation}
where $p_{\bar{\theta}}^y(z|\tau)={p^y_\mathcal{D}(\tau, z)}/{p^y_\mathcal{D}(\tau)}$ and $\bar\theta=\theta$ highlights these distributions have the same parameterization as $p_{{\theta}}^y$ but are wrapped with \texttt{stop\_grad()} operator when calculating gradients for $\theta$~\citep{han2017abp}.
Comparing \cref{eq:rtg,eq:plan}, it is now clear that $z$ plays a similar role as $RTG$ in promoting temporal consistency in autoregressive models. Uniquely, $p_{\bar{\theta}}^y(z|\tau)$ is the temporal abstraction intrinsic to the model, in contrast to step-wise rewards. From a sequential decision-making perspective, $z$ is effectively a \textit{plan} that the agent is persistent to. From a generative modeling perspective, $z$ from different trajectory modes would decompose the density $p^y(a_{t}|s_{0:t}, a_{0:t-1})$, relieving the burden of learning the autoregressive policy $p_{\beta}(a_{t}|s_{0:t}, a_{0:t-1}, z)$. 

One caveat is that the \textit{transition model estimation} should not be conditioned on $y$. Mixing up more trajectory regimes could provide additional regularization for its estimation and generalization. Actually, environment stochasticity is a more concerning issue for autoregressive \textit{behavior cloning}, as highlighted by \citet{yang2022dichotomy, paster2022you, vstrupl2022upside, brandfonbrener2022does, villaflor2022addressing, eysenbach2022imitating}. Among them, \citet{yang2022dichotomy} pinpoints the issue by viewing $RTG$s as deterministic latent trajectory variables, closely related to what we present here. 
Uniquely, the latent variable $z$ in our model is inherently multi-modal (hence very non-deterministic) and ignorant of step-wise rewards. We postulate that the overfitting issue might be mitigated. This is validated by our empirical study inspired by~\citet{paster2022you}. 

Although \textit{RTG prediction} and \textit{plan prediction} both promote temporal consistency, they function very differently when mixing trajectories from multiple return-conditioned regimes. \textit{RTG prediction} is a supervised learning over the joint distribution $\pdata(\tau, \rho)$. Simply mixing trajectories from multiple regimes can't encourage generalization to trajectories that are \textit{stitched} with those in the dataset. \citet{yamagata2023q} propose to resolve this by replacing $RTG$ with $RTG^Q$. Intuitively, this augments the distribution $\pdata(\tau, \rho)$ with $\pdata(\tau', \rho^Q)$, where $\tau'$ denotes trajectories covered by the offline dynamic programming, such as Q learning, and $\rho^Q= (RTG_0^Q, RTG_1^Q, ... Q_H)$. It significantly improves tasks requiring trajectory \textit{stitching}. 
Conversely, \textit{plan prediction} is an unsupervised learning as it samples from $\pdata(\tau, y)p_{\bar\theta}(z|\tau, y)$. As $z$ contains more trajectory-related information than step-wise $RTG$s, trajectories lying outside of $\pdata(\tau, \rho)$ may be in-distribution for $\pdata(\tau, y)p_{\bar\theta}(z|\tau, y)$. The return prediction training further shapes the representation of $z$, which can be benefited from denser coverage of $y$. With more return values covered, we may count on neural networks' strong interpolation capability to shift the trajectory distribution with $y$-conditioning.  
\section{Related work}
\label{sec:literature}
\textbf{Decision-Making via Sequence Modeling} \citet{chen2021decision} propose Decision Transformer (DT), pioneering this paradigm shift. Concurrently, \citet{janner2021offline} explore beam search upon the learned Transformer for model-based planning and inspired later work that searches over the latent state space \citep{zhang2022efficient}. \citet{lee2022multi} report DT's capability in multi-task setting. \citet{zheng2022online} explore the online extension of DT. \citet{yamagata2023q} augment the Monte Carlo RTG in DT with a Q function and show improvement in tasks requiring trajectory \textit{stitching}. \citet{janner2022planning} explore diffusion models~\citep{ho2020denoising} as an alternative generative model family for decision-making. Our model differentiates from all above in data specification and model formulation. 

\textbf{Latent Trajectory Variables in Behavior Cloning} \citet{yang2022dichotomy, paster2022you} investigate the DT's overfitting to environment contingencies and propose latent variable solutions. Our model is closely related to theirs but unique in an EM-style algorithm for MLE. \citet{ajay2021opal, lynch2020learning} propose latent variable models to make Markovian policies temporally extended. Their models are more related to VAE~\citep{kingma2013auto}. 

\textbf{Offline Reinforcement Learning} Since the offline static datasets only partially cover the state transition spaces, efforts from a conventional RL perspective focus on imposing pessimistic biases to value iteration \citep{kumar2020conservative, kostrikov2021offline, uehara2021pessimistic, xie2021bellman, cheng2022adversarially}. \citet{fujimoto2021minimalist} show that simply augmenting value-based methods with behavior cloning achieves impressive performance. \citet{emmons2021rvs} report that supervised learning on return-conditioned policies is competitive to value-based methods in offline RL. Our MLE objective is more related to the supervised learning methods. The latent variable inference further imposes temporal consistency, acting as a replacement of value iteration. 

\textbf{Hierarchical RL} Methods like OPAL~\citep{ajay2021opal}, OPOSM~\citep{freed2023learning} address TD-learning's limitations in long-range credit assignment using a two-stage approach: discovering skills from shorter subsequences to reduce the planning horizon, then applying skill-level CQL or online model-based planning on the reduced horizons. This paper focuses on comparing various methods for long-range credit assignment on the original horizon. Future work includes first discovering skills and then modeling them with a skill-level LPT to further extend the effective horizon.

\section{Experiments}
\label{sec:experiments}
The data specification of trajectory-return pairs distinguishes our empirical study from most existing works in offline RL. Omitting step-wise rewards naturally increases the challenges in decision-making.

\subsection{Overview}
Our empirical study adopts the convention from offline RL. We first train our model with the offline data and then test it as an agent in the corresponding task. More training details and ablation studies of LPT can be found in~\cref{appen:training,appen:ablation}.

\textbf{OpenAI Gym-Mujoco} The D4RL offline RL dataset \citep{fu2020d4rl} features densely-rewarded locomotion tasks including {\em Halfcheetah, Hopper}, and {\em Walker2D}. We test for {\em medium} and {\em medium-replay}. 
It also includes {\em Antmaze}, a locomotion and goal-reaching task with extremely sparse reward. The agent will only receive a reward of 1 if hitting the target location and 0 otherwise. We use its {\em umaze} and {\em umaze-diverse} variants. 

\textbf{Franka Kitchen} 
Franka Kitchen is a multitask environment where a Franka robot with nine degrees of freedom operates within a kitchen setting, interacting with household objects to achieve specific configurations. Our experiments focus on two datasets of the environment: {\em mixed}, and {\em partial}, which consists of non-task-directed demonstrations and partially task-directed demonstrations respectively.

\textbf{Maze2D} Maze2D is a navigation task in which the agent reaches a fixed goal location from random starting positions. The agent is rewarded 1 point when it is around the goal. Experiments are conducted on three layouts: {\em umaze, medium}, and {\em large}, with increasing complexity. The training data of the Maze2D task contains only suboptimal trajectories from and to randomly selected locations. 

\textbf{Connect Four} This is a tile-based game, where the agent plays against a stochastic opponent~\citep{paster2022you}, receiving at the end of an episode 1 reward for winning, 0 for a draw, and -1 for losing. 

\textbf{Baselines} We compare the performance of LPT with several representative baselines including CQL ~\citep{kumar2020conservative}, DT ~\citep{chen2021decision} and QDT ~\citep{yamagata2023q}. CQL baseline results are obtained from \citet{kumar2020conservative}. QDT baseline results are from \citet{yamagata2023q}. The DT results for Gym-Mujoco and Maze2D tasks are from \citet{yamagata2023q}, Antmaze from \cite{zheng2022online}, and Kitchen implemented based on the published source code. CQL and DT results in the Connect Four experiments are from \citet{paster2022you}. The mean and standard deviation of our model, shown as LPT and LPT-EI, are reported over 5 seeds.

\begin{table*}[ht]
\caption{Evaluation results of offline OpenAI Gym MuJoCo tasks. We provide results for data specification with step-wise reward (left) and final return (right). \textbf{Bold} highlighting indicates top scores. LPT outperforms all final-return baselines and most step-wise-reward baselines.}
\label{table:offline}
\resizebox{\linewidth}{!}{%
\centering
\begin{tabular}{lccc|ccccc}
\toprule
\multicolumn{1}{c}{\multirow{2}{*}{Dataset}} & \multicolumn{3}{c}{Step-wise Reward}                                          & \multicolumn{5}{c}{Final Return}        \\
\multicolumn{1}{c}{}                                          & \multicolumn{1}{c}{CQL} & \multicolumn{1}{c}{DT} & \multicolumn{1}{c}{QDT} & \multicolumn{1}{c}{CQL}  & \multicolumn{1}{c}{DT} & \multicolumn{1}{c}{QDT} &\multicolumn{1}{c}{LPT (Ours)} &\multicolumn{1}{c}{LPT-EI (Ours)}\\
\midrule
halfcheetah-medium   &$\bm{44.4}$  &${42.1}$&$42.3$  &$1.0$ &$42.4$ &${42.4}$ &{$43.13\pm0.38$}&{$\bm{43.53}\pm0.08$} \\
halfcheetah-medium-replay  &$\bm{46.2}$ &$34.1$&$35.6$  &$7.8$ &$33.0$ &$32.8$& {$39.64\pm0.83$} & {$\bm{40.66}\pm 0.12$} \\
hopper-medium  &$58.0$    &$60.3$&$66.5$  &$23.3$ &$57.3$ &$50.7$ &{$58.52\pm1.92$} &{$\bm{63.83}\pm 1.47$} \\
hopper-medium-replay    &$48.6$  &$63.7$&$52.1$  &$7.7$ &  $50.8$ &$38.7$ &{$82.29\pm1.26$}&{$\bm{89.93} \pm 0.61$}\\
walker2d-medium  &$79.2$  &$73.3$&$67.1$  &$0.0$ &$69.9$ &$63.7$ &{$77.85\pm3.18$}&{$\bm{81.15} \pm 0.33$}\\
walker2d-medium-replay   &$26.7$  &$60.2$&$58.2$  &$3.2$ & $51.6$ &$29.6$ &{$72.31\pm1.92$} &{$\bm{75.68} \pm 0.34$} \\
\midrule
kitchen-mixed &$51.0$ &$22.3$ &- &- &$17.2$ & - & $61.9 \pm 1.22$ & {$\bm{64.7} \pm 0.51$}\\
kitchen-partial &$49.8$ & $20.4$&- &- &$10.5$ &-&$61.2 \pm 1.75$ & {$\bm{65.3} \pm 0.62$}\\
\bottomrule
\end{tabular}
}
\end{table*}

\subsection{Credit assignment}
When resolving the temporal consistency issue, our model doesn't have an explicit credit assignment mechanism that accounts for the actual contribution of each step. It is not aware of the step-wise rewards either. We are therefore curious about whether the inferred latent variable $z$ can effectively assign fair credits to resolve compounding errors. 
\paragraph{Distributing sparse rewards to high-dimensional actions} 
The Gym-Mujoco environment was a standard testbed for high-dimensional continuous control during the development of modern RL algorithms~\citep{lillicrap2015continuous}. In this environment, step-wise rewards were believed to be critical for TD learning methods. In the setup of offline RL, \citet{chen2021decision} reported the failure of the competitive CQL baseline when delaying step-wise rewards until the end of the trajectories. DT and QDT are reported to be robust to this alternation. As shown in Table~\ref{table:offline}, the proposed model, LPT, outperforms these baselines when the data specifications are the same. Notably, LPT even excels in most of the control tasks when compared with the baselines with step-wise rewards. 

\paragraph{Distributing delayed rewards to long-range sequences} Maze navigation tasks with fully delayed rewards align with our intuition of a planning problem, for it involves decision-making at certain critical states absent of instantaneous feedback. An ideal planner would take in the expected total return and calculate the sequential decisions, automatically distributing credits from the extremely sparse and fully delayed rewards. 
According to \citet{yamagata2023q}, DT fails in these tasks. Our proposed model LPT outperforms QDT by a large margin in all three variants of the maze task. 
These results validate our hypothesis that the additional plan prediction KL imposes temporal consistency on autoregressive policies.  

\begin{table}[t]\label{table:maze2d}
\centering
\caption{Evaluation results of Maze2D tasks. \textbf{Bold} highlighting indicates top scores.}
\resizebox{0.8\linewidth}{!}{%
\centering
\begin{tabular}{lccccc}
\toprule
{Dataset}  &\multicolumn{1}{c}{CQL} & \multicolumn{1}{c}{DT} & \multicolumn{1}{c}{QDT} &\multicolumn{1}{c}{LPT} &\multicolumn{1}{c}{LPT-EI}\\
\midrule
Maze2D-umaze       &{$5.7$} &$31.0\pm21.3$ &$57.3\pm8.2$ & $65.43\pm2.91$ & $\bm{70.57}\pm1.39$\\
Maze2D-medium      &{$5.0$} &$8.2\pm4.4$ &$13.3\pm5.6$ & $20.62\pm1.81$ & $\bm{26.66}\pm0.74$ \\
Maze2D-large   &$12.5$ &$2.3\pm0.9$ &{$31.0\pm19.8$} &{$37.21\pm2.05$}& $\bm{45.89} \pm 2.98$ \\
\bottomrule
\end{tabular}
}
\end{table}

\begin{table}[h]
\label{table:antmaze}
\centering
\caption{Evaluation results of Antmaze tasks. \textbf{Bold} highlighting indicates top scores.}
\resizebox{0.7\linewidth}{!}{%
\centering
\begin{tabular}{lcccc}
\toprule
\multicolumn{1}{l}{Dataset}     &CQL &DT    &LPT & LPT-EI\\ \midrule
Antmaze-umaze  &{${74.0}$} &$53.3\pm5.52$   &{$80.8\pm4.83$} &{$\bm{92.4}\pm0.80$} \\
Antmaze-umaze-diverse  &{${84.0}$}   &$52.5\pm5.89$   &{$78.5\pm1.66$} &{$\bm{84.4}\pm1.96$}\\
\bottomrule
\end{tabular}
}
\end{table}

\begin{figure}[ht]
   \centering
   \begin{minipage}[b]{.42\textwidth}
  \centering
  \includegraphics[width=\textwidth]{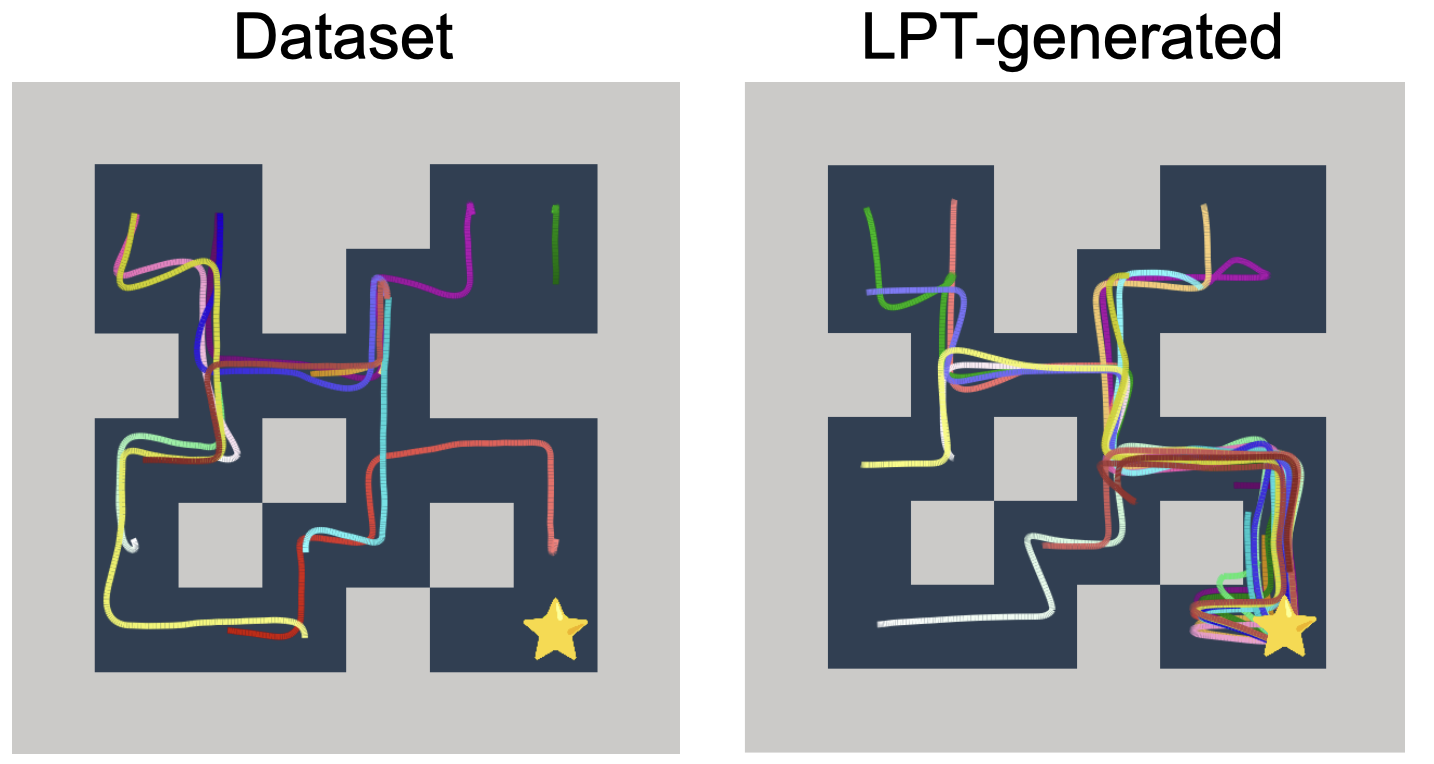}
    \\
   {\small (a) Maze2d-Medium}
   \end{minipage}
   \begin{minipage}[b]{.57\textwidth}
  \centering         
   \includegraphics[width=\textwidth]{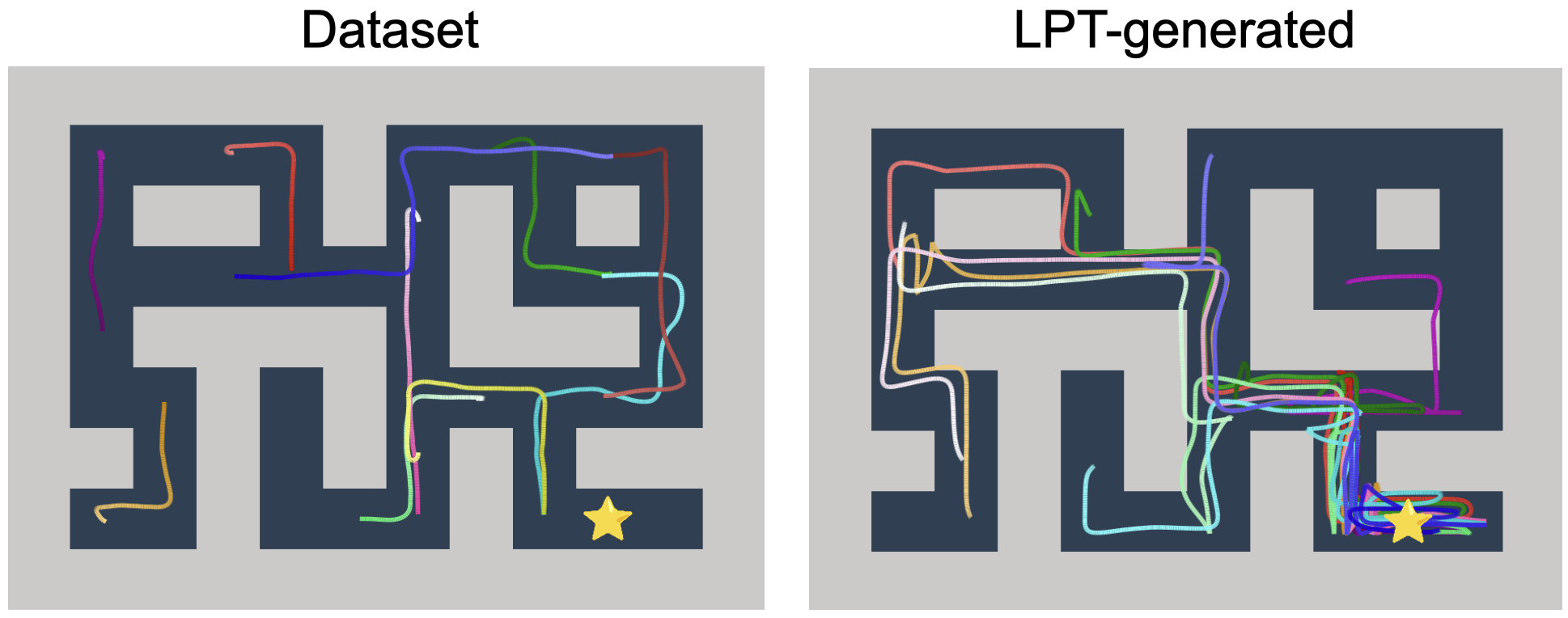}
   \\
   {\small (b) Maze2d-Large }
   \end{minipage}
   \caption{\small (a) Maze2D-medium environment (b) Maze2D-large environment. Left panels show example trajectories from the training set and right panels show LPT generations. Yellow stars represent the goal states. 
   }
    \label{fig:maze2d_traj}
\end{figure}

\subsection{Trajectory \textit{stitching}} 

In addition to credit assignment, the setup of offline RL further presents a challenge, trajectory \textit{stitching}~\citep{fu2020d4rl}, which articulates the problem of shifting the trajectory distribution towards sparsely covered regimes with higher returns. In the Franka Kitchen environment, both the {\em mixed}, and {\em partial} datasets contain undirected data where the robot executes subtasks that do not necessarily achieve the goal configuration. The "mixed" dataset contains no complete solution trajectories, necessitating that the agent learn to piece together relevant sub-trajectories. A similar setting happens in Maze2D domain. Taking Maze2D-medium as an example, in the training set, the average return of all trajectories is $3.98$ with a standard deviation of $10.44$, where the max return is $47$. DT's score is only marginally above the average return. \citet{yamagata2023q} attribute DT's failure in Maze2D to its difficulty with trajectory \textit{stitching}.

\begin{wrapfigure}{l}{0.5\textwidth}
    \vspace{-1.2em}
    \centering
    \includegraphics[width=\linewidth]{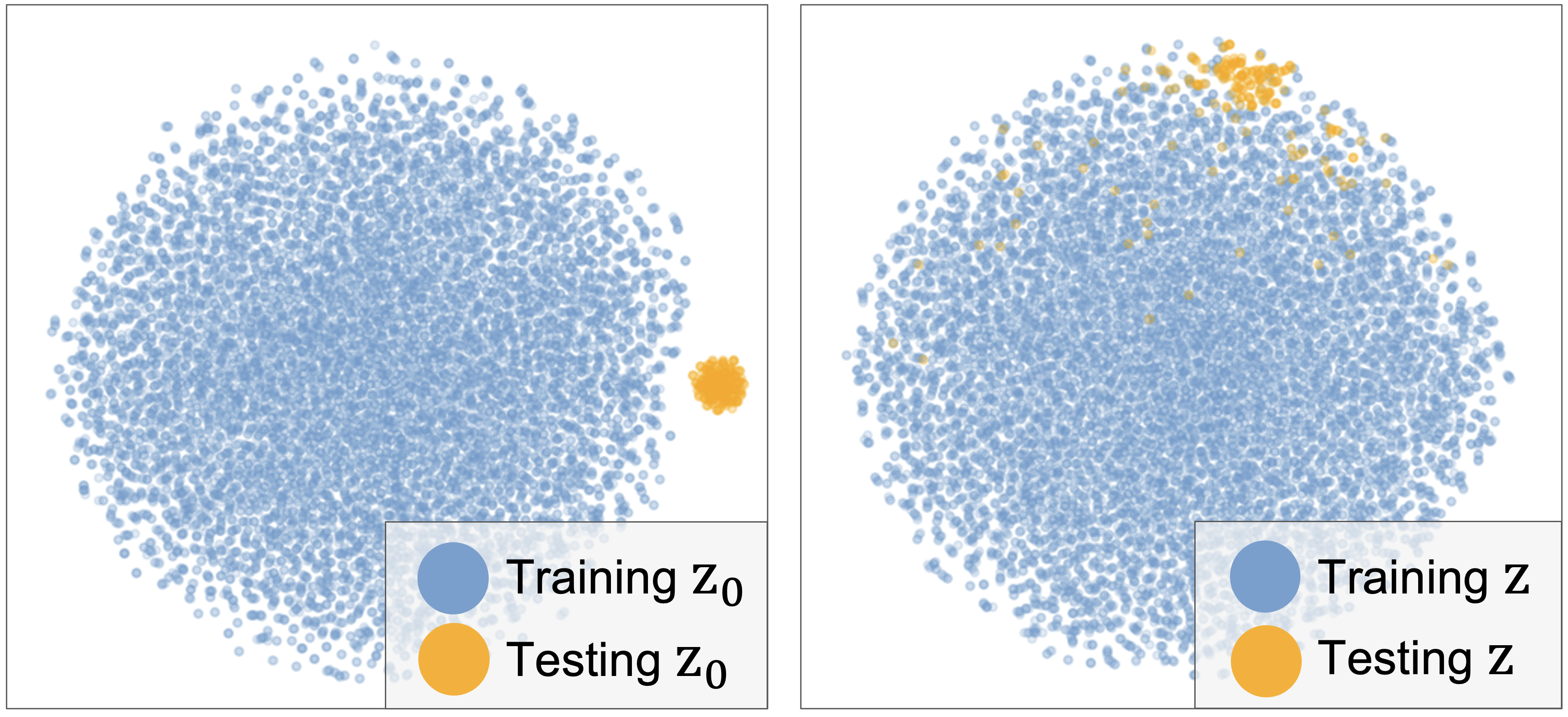}
    \caption{t-SNE plot of latent variables in the Maze2D-medium. Left: Training $z_0$ from aggregated posterior $\E_{p_\mathcal{D}(\tau, y)}[p_\theta(z_0|\tau,y)]$. Testing $z_0$ from $p_\theta(z_0|y)$, disjoint from training population. Right: Distribution of $z=U_\alpha(z_0)$. }
    \label{fig:latent-tsne}
    \vspace{-2em}
\end{wrapfigure}

\cref{fig:maze2d_traj} visualizes samples from the training data and successful trajectories in testing. The left panels show that trajectories in training are suboptimal in terms of (1) being short in length and (2) containing very few goal-reaching instances. Trajectories on the right are generated by $10$ random runs with LPT, where the agent successfully navigates to the end goal from random starting positions in an effective manner. This indicates that the agent can discover the correlation between different $y$s to facilitate such stitching. 

To probe into the agent's understanding of trajectories' returns, we visualize the representation space of the latent variables. The left of \cref{fig:latent-tsne} is the aggregated posterior distribution of $z_0$. We can see that $z_0$ infered from $p_\theta(z_0|y)$ are distant away from the training population. The agent understands they are not very likely in the training set. The right of \cref{fig:latent-tsne} is the distribution of $z$, which is transformed from $z_0$ with the UNet, $z=U_\alpha(z_0)$. We observe that $z$s from the generated trajectories become ``in-distribution'' in the sense that some of them are mingled into the training population and the remaining lie inside a region coverable through linear interpolation of training samples. The agent understands what trajectories to generate even if they are unlikely among what it has seen.

\subsection{Environment contingencies} 

To live in a stochastic world, contingent planning that is adaptable to unforeseen noises is desirable. \citet{paster2022you, yang2022dichotomy} discover that DT's performance would degrade in stochastic environments due to inevitable overfitting towards contingencies. We examine LPT and other baselines in Connect Four from~\citet{paster2022you}. Connect Four is a two-player game, where the opponent will make adversarial moves to deliberately disturb an agent's plan. According to the empirical study from~\citet{paster2022you}, the degradation of DT is more significant than in stochastic Gym tasks from~\citet{yang2022dichotomy}. As shown in Table \ref{table:connect4}, LPT achieves the highest score with minimal variance. The ESPER baseline is from~\citet{paster2022you}, which is very relevant to LPT as it is also a latent variable model. ESPER learns the latent variable model with an adversarial loss. It further adds a clustering loss in the latent space. LPT's on-par performance may justify that MLE upon a more flexible prior can play an equal role. 

\begin{table}[h]
\centering
\caption{Evaluation results on Connect Four. \textbf{Bold} highlighting indicates top scores.}
\label{table:connect4}
\resizebox{0.65\linewidth}{!}{%
\centering
\begin{tabular}{lccccc}
\toprule
{Dataset}  &\multicolumn{1}{c}{CQL} & \multicolumn{1}{c}{DT} &\multicolumn{1}{c}{ESPER} &\multicolumn{1}{c}{LPT}\\
\midrule
Connect Four  &{$0.61\pm0.05$} &$0.8\pm0.07$ &$\mathbf{0.99}\pm0.03$ &$\mathbf{0.99}\pm0.01$\\
\bottomrule
\end{tabular}
}
\end{table} 

\section{Limitation}
\label{sec:discussion}
We omit the Antmaze-large experiment from the main text and included potential reasons for LPT's unsatisfactory performance in~\cref{appen:antmaze}. Another interesting direction is to study LPT's continual learning potential. During planning, LPT explores with provably efficient posterior sampling~\citep{osband2013more, osband2017posterior}.

\section{Summary}
\label{sec:summary}
We study generative modeling for planning in the absence of step-wise rewards. We propose LPT which generates trajectory and return from a latent variable. In learning, posterior sampling of the latent variable naturally gathers sub-trajectories to form an episode-wise abstraction despite finite context in training. In inference, the posterior sampling given the target final return explores the optimal regime of the latent space. It produces a latent variable that guides the autoregressive policy to execute consistently. Across diverse evaluations, LPT demonstrates competitive capacities of nuanced credit assignments, trajectory stitching, and adaptation to environmental contingencies. Contemporary work extends LPT's application to online molecule design~\citep{kong2024molecule}. Future research directions include studying online and multi-agent variants of this model, exploring its application in real-world robotics, and investigating its potential in embodied agents. 

\section*{Acknowledgements}

The work was partially supported by NSF DMS-2015577, NSF DMS-2415226, and a gift fund from Amazon. We sincerely thank Mr. Shanwei Mu and Dr. Jiajun Lu at Akool Research for their computational support, as well as the anonymous reviewers for their valuable feedback.

\newpage
\bibliography{reference}

\begin{thebibliography}{42}
\providecommand{\natexlab}[1]{#1}
\providecommand{\url}[1]{\texttt{#1}}
\expandafter\ifx\csname urlstyle\endcsname\relax
  \providecommand{\doi}[1]{doi: #1}\else
  \providecommand{\doi}{doi: \begingroup \urlstyle{rm}\Url}\fi

\bibitem[Ajay et~al.(2021)Ajay, Kumar, Agrawal, Levine, and Nachum]{ajay2021opal}
Anurag Ajay, Aviral Kumar, Pulkit Agrawal, Sergey Levine, and Ofir Nachum.
\newblock Opal: Offline primitive discovery for accelerating offline reinforcement learning.
\newblock In \emph{International Conference on Learning Representations (ICLR)}, 2021.

\bibitem[Brandfonbrener et~al.(2022)Brandfonbrener, Bietti, Buckman, Laroche, and Bruna]{brandfonbrener2022does}
David Brandfonbrener, Alberto Bietti, Jacob Buckman, Romain Laroche, and Joan Bruna.
\newblock When does return-conditioned supervised learning work for offline reinforcement learning?
\newblock In \emph{Advances in Neural Information Processing Systems (NeurIPS)}, volume~35, pages 1542--1553, 2022.

\bibitem[Chen et~al.(2021)Chen, Lu, Rajeswaran, Lee, Grover, Laskin, Abbeel, Srinivas, and Mordatch]{chen2021decision}
Lili Chen, Kevin Lu, Aravind Rajeswaran, Kimin Lee, Aditya Grover, Misha Laskin, Pieter Abbeel, Aravind Srinivas, and Igor Mordatch.
\newblock Decision transformer: Reinforcement learning via sequence modeling.
\newblock In \emph{Advances in Neural Information Processing Systems (NeurIPS)}, volume~34, pages 15084--15097, 2021.

\bibitem[Cheng et~al.(2022)Cheng, Xie, Jiang, and Agarwal]{cheng2022adversarially}
Ching-An Cheng, Tengyang Xie, Nan Jiang, and Alekh Agarwal.
\newblock Adversarially trained actor critic for offline reinforcement learning.
\newblock In \emph{International Conference on Machine Learning (ICML)}, pages 3852--3878, 2022.

\bibitem[Dhariwal and Nichol(2021)]{dhariwal2021diffusion}
Prafulla Dhariwal and Alexander Nichol.
\newblock Diffusion models beat gans on image synthesis.
\newblock In \emph{Advances in Neural Information Processing Systems (NeurIPS)}, volume~34, pages 8780--8794, 2021.

\bibitem[Emmons et~al.(2021)Emmons, Eysenbach, Kostrikov, and Levine]{emmons2021rvs}
Scott Emmons, Benjamin Eysenbach, Ilya Kostrikov, and Sergey Levine.
\newblock Rvs: What is essential for offline rl via supervised learning?
\newblock \emph{arXiv preprint arXiv:2112.10751}, 2021.

\bibitem[Eysenbach et~al.(2022)Eysenbach, Udatha, Salakhutdinov, and Levine]{eysenbach2022imitating}
Benjamin Eysenbach, Soumith Udatha, Russ~R Salakhutdinov, and Sergey Levine.
\newblock Imitating past successes can be very suboptimal.
\newblock In \emph{Advances in Neural Information Processing Systems (NeurIPS)}, volume~35, pages 6047--6059, 2022.

\bibitem[Freed et~al.(2023)Freed, Venkatraman, Sartoretti, Schneider, and Choset]{freed2023learning}
Benjamin Freed, Siddarth Venkatraman, Guillaume~Adrien Sartoretti, Jeff Schneider, and Howie Choset.
\newblock Learning temporally abstractworld models without online experimentation.
\newblock In \emph{International Conference on Machine Learning (ICML)}, pages 10338--10356, 2023.

\bibitem[Fu et~al.(2020)Fu, Kumar, Nachum, Tucker, and Levine]{fu2020d4rl}
Justin Fu, Aviral Kumar, Ofir Nachum, George Tucker, and Sergey Levine.
\newblock D4rl: Datasets for deep data-driven reinforcement learning.
\newblock \emph{arXiv preprint arXiv:2004.07219}, 2020.

\bibitem[Fujimoto and Gu(2021)]{fujimoto2021minimalist}
Scott Fujimoto and Shixiang~Shane Gu.
\newblock A minimalist approach to offline reinforcement learning.
\newblock In \emph{Advances in Neural Information Processing Systems (NeurIPS)}, volume~34, pages 20132--20145, 2021.

\bibitem[Han et~al.(2017)Han, Lu, Zhu, and Wu]{han2017abp}
Tian Han, Yang Lu, Song{-}Chun Zhu, and Ying~Nian Wu.
\newblock Alternating back-propagation for generator network.
\newblock In \emph{{AAAI} Conference on Artificial Intelligence (AAAI)}, pages 1976--1984, 2017.

\bibitem[Ho and Salimans(2022)]{ho2022classifier}
Jonathan Ho and Tim Salimans.
\newblock Classifier-free diffusion guidance.
\newblock \emph{arXiv preprint arXiv:2207.12598}, 2022.

\bibitem[Ho et~al.(2020)Ho, Jain, and Abbeel]{ho2020denoising}
Jonathan Ho, Ajay Jain, and Pieter Abbeel.
\newblock Denoising diffusion probabilistic models.
\newblock In \emph{Advances in Neural Information Processing Systems (NeurIPS)}, volume~33, pages 6840--6851, 2020.

\bibitem[Janner et~al.(2021)Janner, Li, and Levine]{janner2021offline}
Michael Janner, Qiyang Li, and Sergey Levine.
\newblock Offline reinforcement learning as one big sequence modeling problem.
\newblock In \emph{Advances in Neural Information Processing Systems (NeurIPS)}, volume~34, pages 1273--1286, 2021.

\bibitem[Janner et~al.(2022)Janner, Du, Tenenbaum, and Levine]{janner2022planning}
Michael Janner, Yilun Du, Joshua Tenenbaum, and Sergey Levine.
\newblock Planning with diffusion for flexible behavior synthesis.
\newblock In \emph{International Conference on Machine Learning (ICML)}, pages 9902--9915, 2022.

\bibitem[Kingma and Welling(2014)]{kingma2013auto}
Diederik~P. Kingma and Max Welling.
\newblock Auto-encoding variational bayes.
\newblock In \emph{International Conference on Learning Representations, (ICLR)}, 2014.

\bibitem[Kong et~al.(2024)Kong, Huang, Xie, Honig, Xu, Xue, Lin, Zhou, Zhong, Zheng, and Wu]{kong2024molecule}
Deqian Kong, Yuhao Huang, Jianwen Xie, Edouardo Honig, Ming Xu, Shuanghong Xue, Pei Lin, Sanping Zhou, Sheng Zhong, Nanning Zheng, and Ying~Nian Wu.
\newblock Molecule design by latent prompt transformer.
\newblock In \emph{Advances in Neural Information Processing Systems (NeurIPS)}, 2024.

\bibitem[Kostrikov et~al.(2021)Kostrikov, Nair, and Levine]{kostrikov2021offline}
Ilya Kostrikov, Ashvin Nair, and Sergey Levine.
\newblock Offline reinforcement learning with implicit q-learning.
\newblock In \emph{International Conference on Learning Representations (ICLR)}, 2021.

\bibitem[Kumar et~al.(2020)Kumar, Zhou, Tucker, and Levine]{kumar2020conservative}
Aviral Kumar, Aurick Zhou, George Tucker, and Sergey Levine.
\newblock Conservative q-learning for offline reinforcement learning.
\newblock In \emph{Advances in Neural Information Processing Systems (NeurIPS)}, volume~33, pages 1179--1191, 2020.

\bibitem[Lee et~al.(2022)Lee, Nachum, Yang, Lee, Freeman, Guadarrama, Fischer, Xu, Jang, Michalewski, et~al.]{lee2022multi}
Kuang-Huei Lee, Ofir Nachum, Mengjiao~Sherry Yang, Lisa Lee, Daniel Freeman, Sergio Guadarrama, Ian Fischer, Winnie Xu, Eric Jang, Henryk Michalewski, et~al.
\newblock Multi-game decision transformers.
\newblock In \emph{Advances in Neural Information Processing Systems (NeurIPS)}, volume~35, pages 27921--27936, 2022.

\bibitem[Lillicrap et~al.(2015)Lillicrap, Hunt, Pritzel, Heess, Erez, Tassa, Silver, and Wierstra]{lillicrap2015continuous}
Timothy~P Lillicrap, Jonathan~J Hunt, Alexander Pritzel, Nicolas Heess, Tom Erez, Yuval Tassa, David Silver, and Daan Wierstra.
\newblock Continuous control with deep reinforcement learning.
\newblock \emph{arXiv preprint arXiv:1509.02971}, 2015.

\bibitem[Lynch et~al.(2020)Lynch, Khansari, Xiao, Kumar, Tompson, Levine, and Sermanet]{lynch2020learning}
Corey Lynch, Mohi Khansari, Ted Xiao, Vikash Kumar, Jonathan Tompson, Sergey Levine, and Pierre Sermanet.
\newblock Learning latent plans from play.
\newblock In \emph{Conference on robot learning (CoRL)}, pages 1113--1132, 2020.

\bibitem[Neal(2011)]{neal2011mcmc}
Radford~M Neal.
\newblock {MCMC} using hamiltonian dynamics.
\newblock \emph{Handbook of Markov Chain Monte Carlo}, 2, 2011.

\bibitem[Nijkamp et~al.(2020)Nijkamp, Pang, Han, Zhou, Zhu, and Wu]{nijkamp2020learning2}
Erik Nijkamp, Bo~Pang, Tian Han, Linqi Zhou, Song-Chun Zhu, and Ying~Nian Wu.
\newblock Learning multi-layer latent variable model via variational optimization of short run mcmc for approximate inference.
\newblock In \emph{European Conference on Computer Vision (ECCV)}, pages 361--378, 2020.

\bibitem[Osband and Van~Roy(2017)]{osband2017posterior}
Ian Osband and Benjamin Van~Roy.
\newblock Why is posterior sampling better than optimism for reinforcement learning?
\newblock In \emph{International Conference on Machine Learning (ICML)}, pages 2701--2710, 2017.

\bibitem[Osband et~al.(2013)Osband, Russo, and Van~Roy]{osband2013more}
Ian Osband, Daniel Russo, and Benjamin Van~Roy.
\newblock (more) efficient reinforcement learning via posterior sampling.
\newblock In \emph{Advances in Neural Information Processing Systems (NIPS)}, volume~26, 2013.

\bibitem[Pang et~al.(2020)Pang, Han, Nijkamp, Zhu, and Wu]{pang2020learning}
Bo~Pang, Tian Han, Erik Nijkamp, Song-Chun Zhu, and Ying~Nian Wu.
\newblock Learning latent space energy-based prior model.
\newblock In \emph{Advances in Neural Information Processing Systems (NeurIPS)}, 2020.

\bibitem[Paster et~al.(2022)Paster, McIlraith, and Ba]{paster2022you}
Keiran Paster, Sheila McIlraith, and Jimmy Ba.
\newblock You can’t count on luck: Why decision transformers and rvs fail in stochastic environments.
\newblock In \emph{Advances in Neural Information Processing Systems (NeurIPS)}, volume~35, pages 38966--38979, 2022.

\bibitem[Ronneberger et~al.(2015)Ronneberger, Fischer, and Brox]{ronneberger2015u}
Olaf Ronneberger, Philipp Fischer, and Thomas Brox.
\newblock U-net: Convolutional networks for biomedical image segmentation.
\newblock In \emph{International Conference on Medical Image Computing and Computer-Assisted Intervention (MICCAI)}, pages 234--241, 2015.

\bibitem[Ross and Bagnell(2010)]{ross2010efficient}
St{\'e}phane Ross and Drew Bagnell.
\newblock Efficient reductions for imitation learning.
\newblock In \emph{International Conference on Artificial Intelligence and Statistics (AISTATS)}, pages 661--668, 2010.

\bibitem[{\v{S}}trupl et~al.(2022){\v{S}}trupl, Faccio, Ashley, Schmidhuber, and Srivastava]{vstrupl2022upside}
Miroslav {\v{S}}trupl, Francesco Faccio, Dylan~R Ashley, J{\"u}rgen Schmidhuber, and Rupesh~Kumar Srivastava.
\newblock Upside-down reinforcement learning can diverge in stochastic environments with episodic resets.
\newblock \emph{arXiv preprint arXiv:2205.06595}, 2022.

\bibitem[Sutton and Barto(2018)]{sutton2018reinforcement}
Richard~S Sutton and Andrew~G Barto.
\newblock \emph{Reinforcement learning: An introduction}.
\newblock MIT press, 2018.

\bibitem[Tieleman(2008)]{tieleman2008training}
Tijmen Tieleman.
\newblock Training restricted boltzmann machines using approximations to the likelihood gradient.
\newblock In \emph{International conference on Machine learning (ICML)}, pages 1064--1071, 2008.

\bibitem[Uehara and Sun(2021)]{uehara2021pessimistic}
Masatoshi Uehara and Wen Sun.
\newblock Pessimistic model-based offline reinforcement learning under partial coverage.
\newblock In \emph{International Conference on Learning Representations (ICLR)}, 2021.

\bibitem[Villaflor et~al.(2022)Villaflor, Huang, Pande, Dolan, and Schneider]{villaflor2022addressing}
Adam~R Villaflor, Zhe Huang, Swapnil Pande, John~M Dolan, and Jeff Schneider.
\newblock Addressing optimism bias in sequence modeling for reinforcement learning.
\newblock In \emph{International Conference on Machine Learning (ICML)}, pages 22270--22283, 2022.

\bibitem[Xie et~al.(2016)Xie, Lu, Zhu, and Wu]{xie2016theory}
Jianwen Xie, Yang Lu, Song{-}Chun Zhu, and Ying~Nian Wu.
\newblock A theory of generative convnet.
\newblock In \emph{International Conference on Machine Learning (ICML)}, pages 2635--2644, 2016.

\bibitem[Xie et~al.(2023)Xie, Zhu, Xu, Li, and Li]{XieZXL023}
Jianwen Xie, Yaxuan Zhu, Yifei Xu, Dingcheng Li, and Ping Li.
\newblock A tale of two latent flows: Learning latent space normalizing flow with short-run langevin flow for approximate inference.
\newblock In \emph{{AAAI} Conference on Artificial Intelligence (AAAI)}, pages 10499--10509, 2023.

\bibitem[Xie et~al.(2021)Xie, Cheng, Jiang, Mineiro, and Agarwal]{xie2021bellman}
Tengyang Xie, Ching-An Cheng, Nan Jiang, Paul Mineiro, and Alekh Agarwal.
\newblock Bellman-consistent pessimism for offline reinforcement learning.
\newblock In \emph{Advances in Neural Information Processing Systems (NeurIPS)}, volume~34, pages 6683--6694, 2021.

\bibitem[Yamagata et~al.(2023)Yamagata, Khalil, and Santos-Rodriguez]{yamagata2023q}
Taku Yamagata, Ahmed Khalil, and Raul Santos-Rodriguez.
\newblock Q-learning decision transformer: Leveraging dynamic programming for conditional sequence modelling in offline rl.
\newblock In \emph{International Conference on Machine Learning (ICML)}, pages 38989--39007, 2023.

\bibitem[Yang et~al.(2022)Yang, Schuurmans, Abbeel, and Nachum]{yang2022dichotomy}
Sherry Yang, Dale Schuurmans, Pieter Abbeel, and Ofir Nachum.
\newblock Dichotomy of control: Separating what you can control from what you cannot.
\newblock In \emph{International Conference on Learning Representations (ICLR)}, 2022.

\bibitem[Zhang et~al.(2022)Zhang, Janner, Li, Rockt{\"a}schel, Grefenstette, Tian, et~al.]{zhang2022efficient}
Tianjun Zhang, Michael Janner, Yueying Li, Tim Rockt{\"a}schel, Edward Grefenstette, Yuandong Tian, et~al.
\newblock Efficient planning in a compact latent action space.
\newblock In \emph{International Conference on Learning Representations (ICLR)}, 2022.

\bibitem[Zheng et~al.(2022)Zheng, Zhang, and Grover]{zheng2022online}
Qinqing Zheng, Amy Zhang, and Aditya Grover.
\newblock Online decision transformer.
\newblock In \emph{International Conference on Machine Learning (ICML)}, pages 27042--27059, 2022.

\end{thebibliography}

\appendix

\section{Appendix}
\subsection{Details about model and learning}
\label{appen:model}

Given a trajectory $\tau$, $z \in \mathbb{R}^d$ is the latent vector to represent the variable-length trajectory. $y \in \mathbb{R}$ is the return of the trajectory. With offline training trajectory-return pairs $\{(\tau_i, y_i), i = 1, ..., n\}$. The log-likelihood function is $L( \theta) = \sum_{i=1}^{n} \log p_\theta(\tau_i, y_i)$, with learning gradient $\nabla_\theta L( \theta) = \sum_{i=1}^{n} \nabla_\theta \log p_\theta(\tau_i, y_i)$. We derive the form of $\nabla_\theta \log p_\theta(\tau_i, y_i)$, proving \cref{eq:grad}
 below, dropping index subscript $_i$ for simplicity.
\begin{align*} \nabla_\theta \log p_\theta(\tau, y) &= \frac{\nabla_\theta p_\theta(\tau, y)}{p_\theta(\tau, y)} \\ &= \frac{1}{p_\theta(\tau, y)} \int \nabla_\theta p_\theta(\tau, y, z=U_\alpha(z_0)) dz_0 
\\ &= \int \frac{p_\theta(\tau, y, z=U_\alpha(z_0))}{p_\theta(\tau, y)} \nabla_\theta \log p_\theta(\tau, y, z=U_\alpha(z_0)) dz_0 
\\ &= \int p_\theta(z_0 | \tau, y) \nabla_\theta \log p_\theta(\tau, y, z=U_\alpha(z_0)) dz_0 
\\ &= \E_{p_\theta(z_0|\tau, y)} \left[ \nabla_\theta \log p_\theta(\tau, y, z=U_\alpha(z_0)) \right] 
\\ &= \E_{p_\theta(z_0|\tau, y)} \left[ \nabla_\theta \log p_\beta(\tau|U_\alpha(z_0)) + \nabla_\theta \log p_\gamma(y|U_\alpha(z_0)) + \nabla_\theta \log p_0(z_0) \right] 
\\ &= \E_{p_\theta(z_0|\tau, y)} \left[ \nabla_\theta \log p_\beta(\tau|U_\alpha(z_0)) + \nabla_\theta \log p_\gamma(y|U_\alpha(z_0)) \right]. 
\end{align*}

\subsection{Training details}
\label{appen:training}
For Gym-Mujoco offline training, as shown in Table ~\ref{table:gym-train}, most of the hyperparameters were shared across all tasks except context length and hidden size. However, due to the significant variations in the scale of the maze maps and the lengths of the trajectories within the Maze2D environments—spanning umaze, medium, and large categories—model sizes were adjusted accordingly to accommodate these differences, where the detailed setting can be found in Table ~\ref{table:maze2d-train}. We also show the parameters for Franka Kitchen environment in Table ~\ref{table:kitchen-train} and Connect Four in Table ~\ref{table:connect4-train}. 

Training time for the Gym-Mujoco tasks using a single Nvidia A6000 GPU is 18 hours on average. We train Maze2d tasks using a single Nvidia A100 GPU using 30 hours on average. Kitchen tasks using a single Nvidia A6000 GPU takes 60 hours on average. Connect-4 on a single Nvidia A6000 GPU takes 10 hours. 

\begin{table}[H]
\centering
\caption{Gym-Mujoco Environments LPT Model Parameters}
\centering
\begin{tabular}{lcccc}
\toprule
{Parameter}  &\multicolumn{1}{c}{HalfCheetah} & \multicolumn{1}{c}{Walker2D} &\multicolumn{1}{c}{Hopper}&\multicolumn{1}{c}{AntMaze}\\
\midrule
Number of layers            & 3     & 3     & 3     & 3     \\
Number of attention heads   & 1     &1      &1      & 1     \\
Embedding dimension         &128    &128    &128    & 192     \\
Context length              &32     &64     &64     & 64     \\
Learning rate               &1e-4   &1e-4   &1e-4   & 1e-3     \\
Langevin step size          &0.3    &0.3    &0.3    & 0.3     \\
Nonlinearity function       &ReLU   &ReLU   &ReLU   & ReLU     \\
\bottomrule
\end{tabular}
\label{table:gym-train}
\end{table}

\begin{table}[H]
\centering
\caption{Maze2D Environments LPT Model Parameters}
\centering
\begin{tabular}{lccc}
\toprule
{Parameter}  &\multicolumn{1}{c}{Umaze} & \multicolumn{1}{c}{Medium} &\multicolumn{1}{c}{Large}\\
\midrule
Number of layers            & 1     & 3     & 4     \\
Number of attention heads   & 8     & 1      &4      \\
Embedding dimension         &128    &192    &192    \\
Context length              &32     &64     &64     \\
Learning rate               &1e-3   &1e-3   &2e-4   \\
Langevin step size          &0.3    &0.3    &0.3    \\
Nonlinearity function       &ReLU   &ReLU   &ReLU   \\
\bottomrule
\end{tabular}
\label{table:maze2d-train}
\end{table}

\begin{table}[H]
\centering
\caption{Franka Kitchen Environments LPT Model Parameters}
\centering
\begin{tabular}{lcc}
\toprule
{Parameter}  &\multicolumn{1}{c}{Mixed} & \multicolumn{1}{c}{Partial}\\
\midrule
Number of layers            &4     &3      \\
Number of attention heads   &4     &16      \\
Embedding dimension         &128    &128     \\
Context length              &16     &16      \\
Learning rate               &1e-3   &1e-3    \\
Langevin step size          &0.3    &0.3     \\
Nonlinearity function       &ReLU   &ReLU    \\
\bottomrule
\end{tabular}
\label{table:kitchen-train}
\end{table}

\begin{table}[H]
\centering
\caption{Connect 4 LPT Model Parameters}
\centering
\begin{tabular}{lccc}
\toprule
{Parameter}  &\multicolumn{1}{c}{Value} \\
\midrule
Number of layers            & 3     \\
Number of attention heads   & 4     \\
Embedding dimension         &128    \\
Context length              &4     \\
Learning rate               &1e-3   \\
Langevin step size          &0.3    \\
Nonlinearity function       &ReLU   \\
\bottomrule
\end{tabular}
\label{table:connect4-train}
\end{table}

\subsection{Discussion on data quality of Antmaze medium and large}
\label{appen:antmaze}

\begin{figure}[h]
    \centering
    \includegraphics[width=0.6\textwidth]{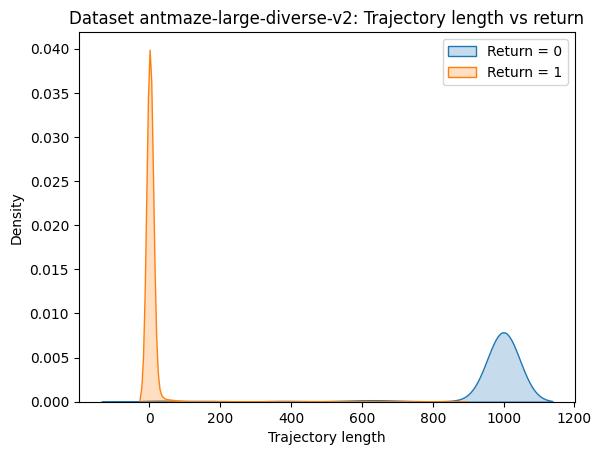}
    \caption{Trajectory length and return distribution in dataset Antmaze-large-diverse}
    \label{fig:len_vs_return}
\end{figure}

\begin{figure}[ht]
    \centering
    \includegraphics[width=0.7\textwidth]{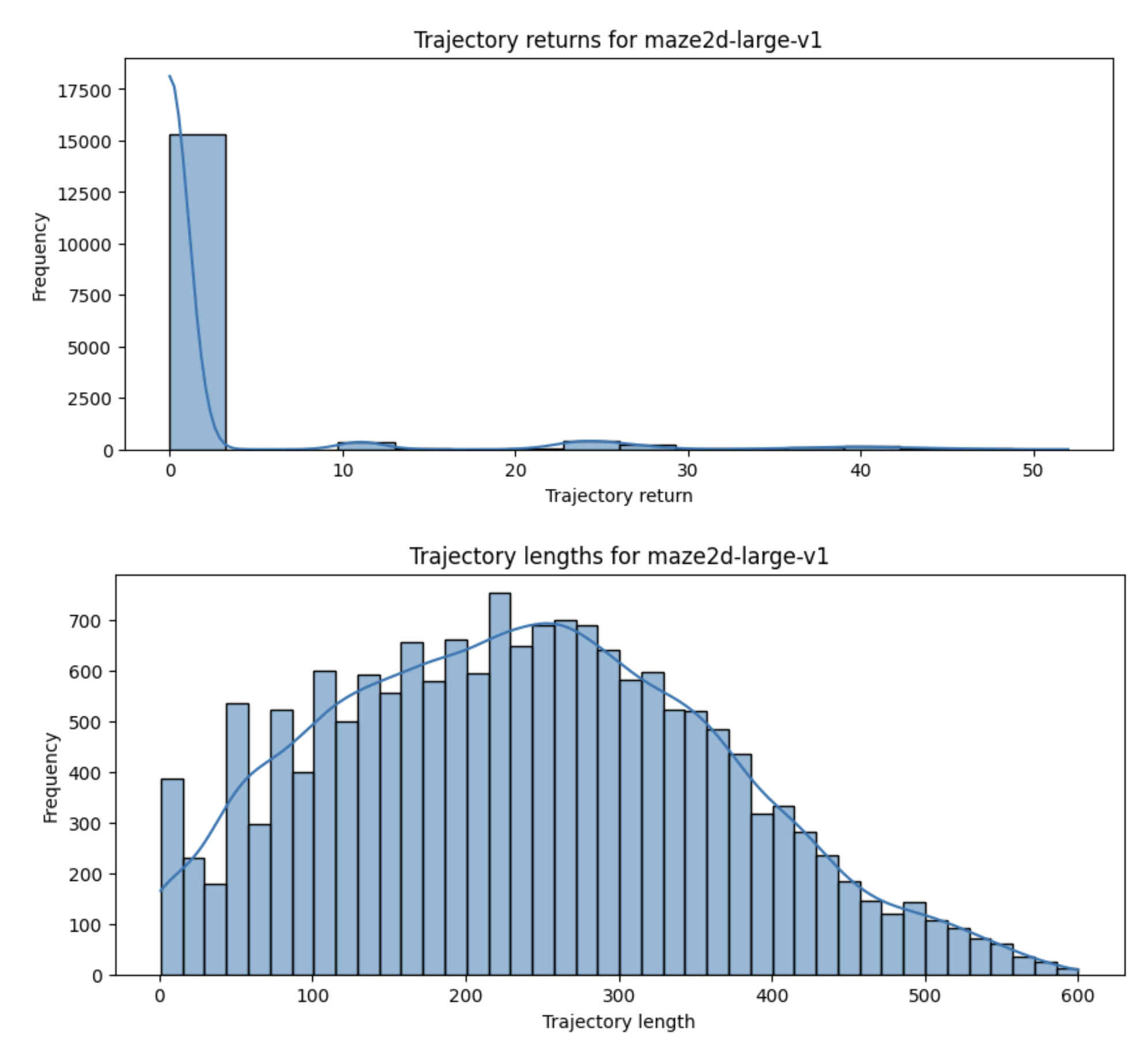}
    \caption{Trajectory length and return distribution in dataset Maze2D-Large}
    \label{fig:stats-large}
\end{figure}

In our experiments, we encounter a curious phenomenon that LPT outperforms CQL, DT and QDT in Antmaze-umaze by a large margin but falls behind in Antmaze-large. Upon closer examination of the data from D4RL, we gained valuable insights into the potential reasons behind LPT's performance on this task.

\cref{fig:len_vs_return} plots the distributions of final returns and the trajectory lengths. Surprisingly, this dataset consists of 5448 trajectories (75.86\%) with length=1, 893 trajectories (12.43\%) with length=1000, and only 841 trajectories (11.71\%) with lengths in between. Such a biased trajectory coverage can be detrimental to sequence models like LPT, which learn to make decisions by discovering correlations between trajectories and returns.

As a reference, \cref{fig:stats-large} shows the distributions of final returns and the trajectory lengths of Maze2D-large, a task where LPT performs well. It is important to note that TD-learning methods, such as CQL and QDT, rely solely on $(s, a, s^\prime, r)$ tuples and are less affected by the trajectory length distribution in the dataset. Consequently, Antmaze-large in D4RL remains a fair dataset for these methods to perform offline RL.

\subsection{Ablation study}
\label{appen:ablation}
We investigate the role of the expressive prior $p_\alpha(z)$ in our Latent Plan Transformer (LPT) model by removing the UNet component, which transforms $z_0$ from a non-informative Gaussian distribution. Table~\ref{table:ablation} reports the results on three Gym-Mujoco tasks and Connect Four. We observe that the performance of LPT drops in all environments when the UNet is removed. For example, in the stochastic environment Connect Four, LPT's performance decreases from $0.99$ to $0.90$, while the baseline Decision Transformer (DT) without latent variables achieves $0.80$. These results indicate that a more flexible prior benefits the learning and inference of LPT.

\begin{table}[h]
\caption{Ablation study results on Gym-Mujoco tasks and Connect Four.}
\label{table:ablation}
\centering
\centering
\begin{tabular}{lcccccc}
\toprule
{Dataset}  &\multicolumn{1}{c}{DT} &\multicolumn{1}{c}{LPT} &\multicolumn{1}{c}{LPT w/o UNet}\\
\midrule
halfcheetah-medium-replay &$33.0\pm 4.8$ & ${39.64}\pm0.83$ &${34.70}\pm1.58$ \\
walker2d-medium-replay &$51.6\pm24.7$ & ${72.31}\pm1.92$ & ${56.88}\pm 4.20$\\
\midrule
Connect Four &$0.8\pm0.07$ &${0.99}\pm0.01$ & ${0.90}\pm0.06$\\
\bottomrule
\end{tabular}
\end{table}

To further explore the impact of the prior, we conducted additional experiments testing the effects of different UNet configurations on LPT's performance. Table~\ref{table:unet_ablation} shows the normalized scores on the \texttt{walker2d-medium-replay} task with various UNet architectures. We observe that reducing the capacity or expressiveness of the UNet (e.g., smaller dimension, fewer multipliers, smaller initial convolution, or fewer ResBlocks) consistently degrades performance, though still outperforming the model without the UNet prior. This suggests that a more expressive prior enhances LPT's ability to model complex policies.

\begin{table}[h]
\caption{Effect of different UNet configurations on LPT performance.}
\label{table:unet_ablation}
\centering
\centering
\begin{tabular}{lc}
\toprule
{Model Prior} & \multicolumn{1}{c}{Normalized Score }\\
\midrule
UNet (original)                  & $\mathbf{72.31} \pm 1.92$ \\
UNet w/ smaller dimension        & $64.06 \pm 1.94$ \\
UNet w/ fewer multipliers        & $64.59 \pm 1.54$ \\
UNet w/ smaller initial convolution & $70.49 \pm 2.84$ \\
UNet w/ single ResBlock          & $67.95 \pm 4.64$ \\
No UNet (Standard Normal prior)  & $56.88 \pm 4.20$ \\
\bottomrule
\end{tabular}
\end{table}

Our results underscore the crucial role of the learned prior in LPT's performance. The original UNet configuration achieves the highest normalized score, indicating that our current UNet design is optimal among the variants tested. We appreciate the reviewer's suggestion, as it prompted us to perform a more detailed analysis of the prior's impact on LPT.


\end{document}